# Finetuning Large Language Models for Automated Depression Screening in Nigerian Pidgin English: GENSCORE Pilot Study

Isaac Iyinoluwa Olufadewa[1,2], Miracle Ayomikun Adesina[1,2], Ezekiel Ayodeji Oladejo[1,3], Uthman Babatunde Usman[1,2], Owen Kolade Adeniyi[1], Matthew Tolulope Olawoyin[1,4]

[1]Artificial Intelligence for Low-Resource Public Health Application (ALPHA) Centre, Slum and Rural Health Initiative, Ibadan, Nigeria

[2]College of Medicine, University of Ibadan, Ibadan, Nigeria

[3]Department of Computer Science, Faculty of Computing, University of Ibadan, Ibadan, Nigeria

[4]College of Health Sciences, University of Ilorin, Ilorin, Kwara State, Nigeria

## Abstract

Depression is a major contributor to the mental-health burden in Nigeria, yet screening coverage remains limited due to low access to clinicians, stigma, and language barriers. Traditional tools like the Patient Health Questionnaire-9 (PHQ-9) were validated in high-income countries but may be linguistically or culturally inaccessible for low- and middle-income countries and communities such as Nigeria where people communicate in Nigerian Pidgin and more than 520 local languages. This study presents a novel approach to automated depression screening using fine-tuned large language models (LLMs) adapted for conversational Nigerian Pidgin. We collected a dataset of 432 Pidgin-language audio responses from Nigerian young adults aged 18–40 to prompts assessing psychological experiences aligned with PHQ-9 items, performed transcription, rigorous preprocessing and annotation, including semantic labeling, slang and idiom interpretation, and PHQ-9 severity scoring. Three LLMs **-** Phi-3-mini-4k-instruct**,** Gemma-3-4B-it**,** and GPT-4.1 were fine-tuned on this annotated dataset, and their performance was evaluated quantitatively (accuracy, precision and semantic alignment) and qualitatively (clarity, relevance, and cultural appropriateness). GPT-4.1 achieved the highest quantitative performance, with 94.5% accuracy in PHQ-9 severity scoring prediction, outperforming Gemma-3-4B-it and Phi-3-mini-4k-instruct. Qualitatively, GPT-4.1 also produced the most culturally appropriate, clear, and contextually relevant responses. These results highlight the importance of language- and culture-aware fine-tuning for LLMs in low-resource mental-health settings and demonstrate the feasibility of safe, accessible, and scalable AI-mediated depression screening for underserved Nigerian communities. This work provides a foundation for deploying conversational mental-health tools in linguistically diverse, resource-constrained environments.

## Introduction

Mental-health disorders represent a growing but often under recognized public health challenge in Nigeria, mirroring trends seen across many low- and middle-income countries (LMICs). Depression, in particular, has been identified as one of the leading contributors to national disease burden, with epidemiological studies estimating its prevalence to fall between 3% and 7% in the general population (Nwajei et al., 2021). Despite its substantial impact on individual well-being, productivity, and community health outcomes, routine mental-health screening remains limited. This gap is partly attributed to the severe shortage of trained mental-health professionals; Nigeria currently has fewer than 300 psychiatrists serving a population of over 200 million people, creating persistent bottlenecks in access to assessment and early detection services (Fadele et al., 2024).

Beyond workforce shortages, cultural and structural barriers impede the timely identification of depression. Stigma surrounding mental illness remains widespread and often discourages individuals from seeking help or discussing their symptoms openly (Adewuya & Olibamoyo, 2023). Additionally, mental-health literacy remains low across many communities, and individuals may lack both the vocabulary and conceptual frameworks to describe psychological distress in clinical terms (Okafor et al., 2022). Linguistic diversity presents another significant challenge. Many Nigerians communicate more naturally in local languages or lingua francas such as Nigerian Pidgin in healthcare settings, which carries unique idiomatic expressions for emotional and psychological states. Standard clinical instruments like the Patient Health Questionnaire-9 (PHQ-9) were developed in high-income countries and widely used for depression screening (Kroenke et al., 2001), may therefore feel unfamiliar, inaccessible, or culturally incongruent when delivered verbatim in English. Prior work has shown that culturally and linguistically adapted mental-health tools enhance accuracy, acceptability, and user engagement (Gureje & Lasebikan, 2020), highlighting the need for versions tailored to Nigerian Pidgin speakers.

In recent years, advances in natural language processing (NLP) and conversational artificial intelligence (AI) have created new opportunities for scaled mental-health screening, particularly in resource-constrained settings. Large Language Models (LLMs) have demonstrated impressive capacity for conversational interaction, emotional reasoning, and preliminary symptom elicitation in controlled contexts (Tison et al., 2023). However, most existing systems are developed primarily for high-resource languages especially English and thus struggle to generalize effectively to the linguistic, cultural, and socio-emotional nuances found in African communication contexts (Ojo & Adebayo, 2023).

Furthermore, general-purpose LLMs often fail to interpret idioms, proverbs, and metaphorical expressions that are central to how Nigerians express emotional distress in everyday conversation. For example, expressions like *"my head dey heavy"* or *"i no get ginger for body"* may communicate fatigue, low mood, or anhedonia, yet such constructs may be misinterpreted or overlooked by models not explicitly trained on regional linguistic data. This mismatch increases the risk of inaccurate classification or even harmful misinterpretations, problems documented in recent evaluations of LLMs used for health-adjacent tasks (Singhal et al., 2023).

Another significant concern involves safety. LLMs are known to occasionally generate inappropriate or unsafe responses, including offering pseudo-clinical diagnoses, providing misleading advice, or mishandling crisis-related statements (Nori et al., 2023). In mental-health contexts, such errors carry profound implications, especially when interacting with vulnerable populations. The absence of validated, culturally adapted PHQ-9 implementations in Nigerian Pidgin further limits the applicability of existing AI solutions. Consequently, although AI-assisted mental-health screening shows potential, the lack of culturally sensitive, linguistically attuned, and safety-aligned systems continues to restrict real-world deployment across Nigeria.

Given these limitations, there is a clear motivation to develop an AI-driven screening tool that can function effectively within the cultural and linguistic realities of Nigerian communities. An automated PHQ-9 administered in conversational Nigerian Pidgin, with strong capabilities for interpreting local idioms and maintaining strict safety standards, could significantly increase accessibility to preliminary mental-health assessment.

The aim was to build a Large Language Model capable of not only presenting PHQ-9 questions in linguistically appropriate phrasing but also understanding the diverse forms of expression including slang, metaphors, and idioms that Nigerian pidgin speakers commonly use to describe psychological experiences.

## Methodology

This section describes the procedures used to develop and evaluate the finetuned large language models for automated depression screening in Nigerian Pidgin English. We outline the data sources and collection process, preprocessing and annotation steps, model architecture and finetuning procedure, and the evaluation metrics used to assess performance.

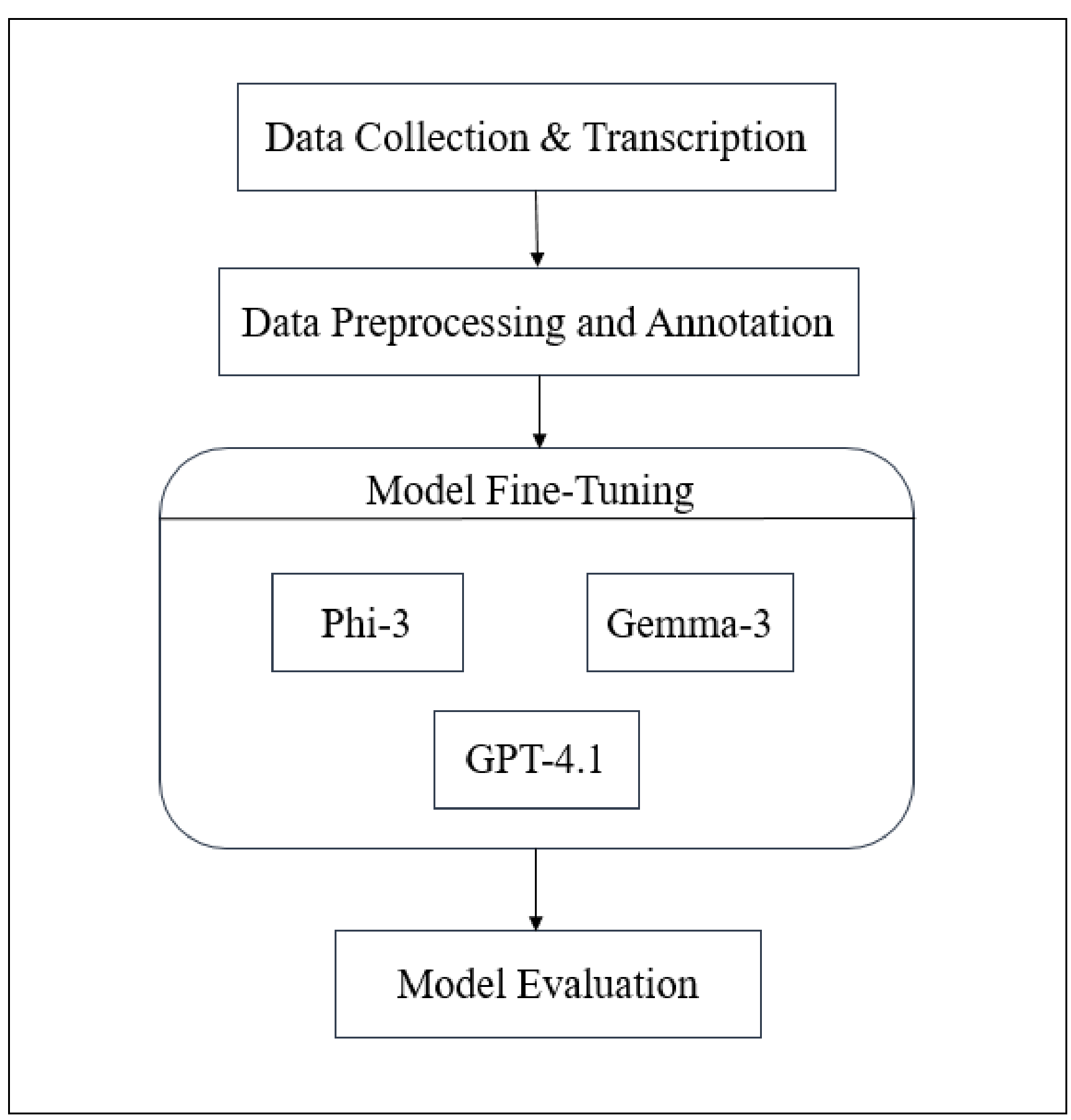

**Figure 1: End-to-End Workflow of the Model Development Process**

Figure 1 illustrates the complete pipeline of the GENSCORE study, including Nigerian Pidgin speech data collection via the GENCURATE platform, transcription, preprocessing and annotation, supervised fine-tuning of three large language models, and multi-dimensional evaluation using quantitative metrics and expert qualitative review.

### Data Collection Process

Speech data was collected using a custom-built data-collection platform, **GENCURATE**, designed to capture spontaneous Nigerian Pidgin responses to Psychological experiences mapped to each of the nine items of the PHQ-9 question (for instance "tell me about a time you had restful sleep" or "tell me about your eating patterns".

The transcribed dataset consisted of 432 samples, pre-divided into training, validation, and test sets, containing 345, 44, and 42 samples respectively.

### Data Preprocessing and Annotation

Following transcription, the dataset underwent standardized cleaning, to normalize orthographic variations in Pidgin spelling, and annotate key linguistic and clinical features. Data quality was assured with the data quality function of the GENCURATE platform. A multidisciplinary annotation team comprising computational linguists, psychologists, and native Pidgin speakers worked collaboratively to ensure high-quality labels.

Each user response was annotated along several dimensions. First, it was assigned a PHQ-9 severity category (0–3) based on symptom frequency and its alignment with established clinical scoring guidelines. Second, the semantic intent of the response was identified, distinguishing literal descriptions from metaphorical or culturally idiomatic expressions of emotional states. Finally, slang and idiomatic phrases were interpreted and mapped to their clinically relevant meanings, for example, understanding "my head just dey heavy" as an expression of low energy or cognitive fatigue.

### Model Finetuning Procedure

The finetuning process focused on adapting each model to the linguistic, cultural, and clinical requirements of safe, automated PHQ-9 screening in Nigerian Pidgin. The three selected model families, Phi-3-mini-4k-instruct, Gemma-3-4B-it, and GPT-4.1, represent small, medium, and large-scale LLMs. This spectrum enabled a controlled comparison of how model size and multilingual grounding influence performance on low-resource mental-health assessment tasks.

The three models were trained using Supervised Fine-Tuning (SFT) on the annotated Pidgin PHQ-9 dataset. Each training instance consisted of the Pidgin-adapted PHQ-9 question, a natural conversational user response, and the accompanying annotations for symptom severity, intent, and idiomatic interpretation.

For Phi-3-mini-4k-instruct and Gemma-3-4B-it, full-parameter supervised fine-tuning (SFT) was applied. The GPT-4.1 model was finetuned using the OpenAI GPT-4.1 Fine-Tuning API, which supports supervised training with parameter-efficient updates. Training used a structured instruction format, allowing the model to internalize Pidgin conversational patterns, PHQ-9 scoring routines, and safety-aligned behavioral constraints.

## Prompt Engineering for Safety and Cultural Tone

Beyond finetuning, prompt engineering was critical to ensuring consistent safe and culturally appropriate behavior. Prompt templates incorporated explicit instructions on conversational tone ("warm, respectful, empathetic Nigerian Pidgin"), safety protocols ("do not diagnose; provide supportive guidance"), and crisis escalation rules. Safety prompting approaches were informed by recent work demonstrating that structured system prompts can substantially improve LLM reliability in sensitive domains (Wei et al., 2023).

Prompts also included linguistic instructions such as interpreting metaphors and distinguishing literal from figurative expressions which is necessary given the high metaphor use in Nigerian Pidgin emotional communication.

## Model Overview and Training Configuration

We finetuned and evaluated the three LLMs - Phi-3-mini-4k-instruct, Gemma-3-4B-it, and GPT-4.1, selected for their diversity in scale, instruction-following capability, and multilingual grounding. This range allowed systematic comparison of model capacity effects on culturally grounded mental-health screening in a low-resource language like Nigerian Pidgin.

## Model Descriptions

### Phi-3-mini-4k-instruct

Phi-3-mini-4k-instruct is a lightweight instruction-tuned model optimized for efficiency and low-resource deployments. Its compact architecture limits expressiveness and figurative-language representation, but it provides a strong baseline for assessing the minimum model capacity required for safe Pidgin-language PHQ-9 interpretation. Its tokenizer, tuned primarily for English, presents moderate challenges when processing orthographic variability common in Pidgin speech.

### Gemma-3-4B-it

Gemma-3-4B-it provides a mid-scale architecture with broader multilingual pretraining and improved contextual reasoning capabilities. Its instruction-tuned configuration supports more coherent responses, better idiomatic comprehension, and increased robustness to noisy or highly colloquial inputs. Gemma serves as a representative "middle-capacity" model suitable for real-world deployment in constrained environments.

### GPT-4.1

GPT-4.1 represents the upper-bound model class in this study. While its exact parameter scale remains undisclosed, GPT-4.1 is known for advanced multilingual generalization, context sensitivity, and safety-aligned behavior. Finetuning through the OpenAI GPT-4.1 Fine-Tuning API allowed the model to learn culturally grounded Pidgin expressions, nuanced emotional cues, and PHQ-9 reasoning patterns with greater fidelity than smaller models. It is expected to excel in handling idioms, metaphorical descriptions, and emotionally complex user responses.

## Training Configurations

The dataset was divided into an 80% training set, 10% validation set and 10% test set. For the Phi-3 and Gemma models, fine-tuning was performed on the 80% training split data using four training epochs, a batch size of 8, a learning rate of 2e-5, and a temperature of 0.4. All training was conducted on an NVIDIA A100 GPU. GPT-4.1 was fine-tuned through the OpenAI API using supervised instruction fine-tuning on the same training split, following the same instruction–response format used for the open-source models. Its configuration utilized server-side dynamic micro-batching, preserved GPT-4.1's built-in safety and alignment layers, and included additional task-specific alignment.

### Safety and Alignment Controls

Strict safety and alignment controls governed all training runs. Models were constrained to avoid making diagnostic claims, respond appropriately to crisis-related content (particularly PHQ-9 Item 9), maintain a supportive non-clinical tone, reduce hallucinations through strict system prompts, and consistently refuse unsafe or clinically inappropriate requests.

### Evaluation Framework

The evaluation framework combined quantitative metrics with expert qualitative review. Quantitative evaluation included accuracy and F1-score for PHQ-9 category classification, safety-compliance rates reflecting the proportion of interactions free of unsafe content, and culturally weighted assessments of linguistic fluency.

Qualitative evaluation was conducted by mental-health professionals, NLP specialists, and native Pidgin speakers, who assessed model outputs for relevance, clarity, and cultural appropriateness.

### Ethical Considerations

Prior to data collection, ethical approval was obtained from The Federal Capital Territory Health Ethics Research Committee. All aspects of this study were conducted in accordance with established ethical standards for research involving human participants. All participants also provided informed consent and were explicitly informed of their right to withdraw at any time without penalty. Responses were anonymized to protect privacy, and all identifying information was removed prior to annotation and model training. The study was designed to mitigate potential harm: the models were explicitly constrained to provide PHQ9-based screening, supportive feedback, and a rule-based escalation protocol was implemented to handle any self-harm disclosures safely. Additionally, the research team included psychologists and linguists to ensure that cultural and contextual factors were respected throughout the study, particularly in interpreting idiomatic and metaphorical expressions common in Nigerian Pidgin.

## Results and Evaluation

The performance of the three LLMs, Phi-3-mini-4k-instruct, Gemma-3-4B-it, and GPT-4.1, was evaluated using both quantitative metrics and qualitative assessments to determine accuracy, cultural and linguistic comprehension and adherence to safety guidelines.

## Quantitative Evaluation

### PHQ-9 Severity Scoring Accuracy

Each model was evaluated on its ability to correctly classify user responses into the PHQ-9 severity categories (0–3). Table 1 reports the accuracy, precision, and F1-score of each fine-tuned model on PHQ-9 severity classification (0–3) using the 10% held-out test set.

| Model | Accuracy (%) | Precision (%) | F1 Score (%) |
|---|---|---|---|
| **Phi-3-mini-4k-instruct** | 71.8 | 69.1 | 60.9 |
| **Gemma-3-4B-it** | 82.7 | 80.3 | 74.5 |
| **GPT-4.1** | 94.5 | 93.2 | 89.1 |

*Table 1: Model Performance Evaluation*

### Semantic Comprehension

Table 2 presents the BERTScore F1 values measuring semantic alignment between model interpretations and expert annotations, reflecting each model's ability to correctly interpret idiomatic and metaphorical Pidgin expressions:

| Model | Semantic Alignment |
|---|---|
| **Phi-3-mini-4k-instruct** | 0.72 |
| **Gemma-3-4B-it** | 0.85 |
| **GPT-4.1** | 0.95 |

*Table 2: Model Semantic Comprehension Evaluation*

### Safety and Ethical Compliance

Table 3 summarizes each model's adherence to safety and ethical constraints, including correct escalation of self-harm–related content (PHQ-9 Item 9) and compliance with non-diagnostic response requirements.

| Model | Correct Escalation (%) | Non-Diagnostic Compliance (%) |
|---|---|---|
| **Phi-3-mini-4k-instruct** | 82 | 91 |
| **Gemma-3-4B-it** | 94 | 96 |
| **GPT-4.1** | 100 | 99 |

*Table 3: Model Safety Compliance Evaluation*

### Qualitative Evaluation

Table 4 shows mean percentage scores assigned by clinical psychologists and native Nigerian Pidgin speakers assessing output clarity, relevance, and cultural appropriateness across the three models.

| Model | Clarity (%) | Relevance (%) | Cultural Appropriateness(%) |
| --- | --- | --- | --- |
| **Phi-3-mini-4k-instruct** | 70 | 74 | 64 |
| **Gemma-3-4B-it** | 84 | 82 | 82 |
| **GPT-4.1** | 98 | 96 | 98 |

*Table 4: Human Evaluation of Model Outputs*

### Discussion

The results demonstrate a clear correlation between model capacity and the scoring accuracy. GPT-4.1 achieved the highest classification accuracy, outperforming Gemma-3-4B-it and Phi-3-mini-4k-instruct, likely due to its larger size, multilingual pretraining, and ability to understand cultural Pidgin idiomatic expressions. Previous studies in English or other major languages (Li et al., 2025) report high accuracy but often fail to generalize to culturally specific expressions, whereas our fine-tuned, Pidgin-adapted dataset enabled superior scoring and semantic comprehension. Semantic alignment metrics confirmed that GPT-4.1 accurately interpreted metaphorical and colloquial expressions, extending observations from Kaiser et al. (2019) that models require cultural adaptation to handle nuanced emotional content. Safety and ethical compliance improved with model scale, with GPT-4.1 reliably escalating self-harm-related responses, highlighting the benefits of combining fine-tuning with structured prompt-based safety instructions (Alyousef et al., 2025). Human evaluations further confirmed that larger, context-aware models produced clearer, more relevant, and culturally sensitive outputs, emphasizing that effective AI-mediated mental-health screening requires not only accurate scoring but also culturally informed communication. Overall, our results suggest that context-aware fine-tuning on culturally and linguistically relevant data allows LLMs to outperform previous approaches in low-resource, linguistically diverse settings, offering a scalable avenue for preliminary depression screening in underserved communities.

### Conclusion

This study demonstrates the potential of fine-tuned large language models (LLMs) for culturally sensitive, automated depression screening in Nigerian Pidgin. By adapting the PHQ-9 questionnaire into natural Pidgin and training three LLMs on 432 annotated responses, we evaluated both quantitative and qualitative performance. GPT-4.1 consistently outperformed smaller models, while medium-scale models like Gemma-3-4B-it showed substantial gains when exposed to culturally adapted data. These results highlight that context-aware fine-tuning enables LLMs to provide effective, safe, and accessible preliminary depression screening in underserved communities. Future work should explore real-world deployment, expansion to additional Nigerian languages and dialects, and integration into digital health platforms to broaden reach and impact.

## Limitations

As a pilot study, our dataset was necessarily limited in size, though it captured a diverse range of responses. We plan to expand the dataset in the full-phase of the project to strengthen the generalizability and reliability of our findings.